# CAN WE FALL IN LOVE WITH AI FICTION?

## THE AI-FICTION PARADOX

**Katherine Elkins**

Integrated Program in Humane Studies, Kenyon College

AI CoLab, Kenyon College

Human Centered AI Lab

**ABSTRACT**

AI development has a fiction dependency problem. Developers have treated large corpora of modern books, including fiction, as valuable enough to accept substantial cost and legal risk, yet current models still struggle to generate compelling long-form fiction. I term this the "AI-Fiction Paradox," and it is particularly startling because training data strongly shapes model output. This paper offers a theoretically precise account of why fiction resists AI generation by identifying three distinct challenges for current systems. First, fiction depends on what I call narrative causation, a form of plot logic where events must feel both surprising in the moment and retrospectively inevitable. Standard autoregressive generation commits to prose sequentially, creating a practical obstacle to coordinating local surprise with retrospective inevitability across a long narrative. Second, I identify an informational revaluation challenge: fiction repeatedly requires the significance of earlier details to be reinterpreted in light of later developments, a form of long-range reasoning that current systems perform unreliably. Third, drawing on over seven years of collaborative research on sentiment arcs, I argue that fiction that moves us requires multi-scale emotional architecture, the orchestration of sentiment at word, sentence, scene, and arc levels simultaneously. Together, these three challenges help explain both why developers have sought large modern book corpora and why compelling long-form fiction remains so difficult to replicate. The analysis also raises urgent questions about what happens when these challenges are overcome. Fiction concentrates unusually powerful cognitive and emotional patterns for modeling human behavior, and mastery of these patterns by AI systems would represent not just a creative achievement but a potent vehicle for human manipulation at scale.

## AI's Modern Fiction Dependency Problem

AI development has a fiction dependency problem: developers have treated modern books, including fiction, as valuable training data, yet current models still struggle to generate compelling long-form fiction. I call this the "AI-Fiction Paradox," and it is striking because training data typically strongly shapes model output, and yet, in the case of fiction, it does not. With thousands of novels represented in major book corpora, we might expect far better long-form fiction results.

The AI-Fiction Paradox thus raises several key questions. Why is modern fiction so important for AI development, even driving companies to risk legal battles for pirated novels? Given how much fiction appears in the training data, why is AI still not very good at generating it? What are the specific limitations in generating fiction, and what do these weaknesses suggest about modern fiction that distinguishes it from other text genres that are more easily generated? Answering these questions will bring us closer to determining whether AI models will ever generate quality fiction that we could fall in love with. But it also raises new questions about fiction as a powerful emotional technology, one that, if it were mastered by AI, would offer a vehicle for human manipulation.

In the following pages, I explore ways in which current failures may be tied to how models are built and deployed rather than to an absence of fictional training data. These struggles highlight properties of fiction that may explain why book corpora are valuable for modeling human behavior. Still, AI limitations are unlikely to be permanent, and if fiction concentrates patterns of human behavior that remain difficult to acquire elsewhere, scholars of fiction should consider whether building better fiction-generating machines might open doors best left closed.

Let's begin, then, by being precise. AI actually has a *modern* fiction dependency problem. I say modern because there is plenty of fiction training data that is out of copyright and widely available on Gutenberg, fiction like *Robinson Crusoe* and *Jane Eyre*. AI developers have nevertheless treated modern book corpora as valuable enough to accept significant legal risk.[1]

---

[1] The two leading cases have produced different and still-evolving results. In *Bartz v. Anthropic*, the court held that using lawfully acquired books for model training could constitute fair use but declined, on the summary-judgment record, to treat Anthropic's acquisition and retention of pirated copies as fair use. Anthropic later agreed to a proposed $1.5 billion class settlement covering nearly half a million works. Final approval remained pending as of July 2026. In *Kadrey v. Meta*, the court granted Meta partial summary judgment on the named plaintiffs' training-copy claim while stressing the limits of the record before it. Separate claims concerning torrenting and alleged distribution remained in litigation as of July 2026. The cases therefore distinguish, at least provisionally, between downstream training use and the means by which copies were acquired or distributed.

Meta trained Llama models on book corpora that included Books3 and material obtained from LibGen even though internal communications described the LibGen proposal as "medium-high legal risk." One internal exchange even characterized books as "actually more important than web data" for training (*Kadrey v. Meta*; Reisner, "Unbelievable Scale"). To be clear, Books3 is not simply a fiction dataset, and Reisner estimates that approximately two-thirds of its titles are nonfiction, leaving roughly 60,000 works of fiction. Even so, this corpus represented a major expansion beyond earlier literary sources such as Project Gutenberg. The scale and internal communications make clear that books were not a marginal data source, although the record does not establish which genres or textual properties produced the value developers sought.

One element of the AI-Fiction Paradox, however, is still rather surprising. Books3 contains not only the contemporary literary fiction most typically found on a college syllabus—works often prized for formal innovation, narrative complexity, or social critique—but also a vast range of popular fiction that scholars typically value far less. The dataset, in other words, includes critically acclaimed literary fiction alongside romance novels, Nobel Prize-winning works alongside self-published erotica, experimental postcolonial narratives alongside licensed fantasy tie-in novels set in gaming worlds like Dungeons & Dragons' Forgotten Realms.[2] The inclusion of both literary and popular fiction suggests that corpus construction was therefore decidedly not guided primarily by conventional judgments of literary merit. So why would an AI company accept legal risk requiring senior-level approval for such an eclectic collection unless scale, variety, or some broadly shared properties of books mattered?[3]

---

[2] See Alex Reisner's *Atlantic* article for a detailed account of the Books3 collection.

The legal risk Meta ran underscores the seriousness of AI's book dependency. Documents produced in *Kadrey* reveal engineers hesitating to torrent massive book collections, with one even objecting, "torrenting from a corporate laptop doesn't feel right." Other internal records, including material escalated to executives and ultimately approved by "MZ," describe the project as a "medium-high legal risk" and discuss stripping copyright markers and other metadata from the texts (Reisner, "Unbelievable Scale").[4] OpenAI's GPT-3 training mixture likewise assigned an 8% sampling weight to each of two undisclosed book corpora, Books1 and Books2, even though they contained far fewer tokens than Common Crawl (Brown et al.). Their contents and provenance were not publicly documented. Together, these choices suggest that large book corpora were worth senior-level attention and potential litigation.[5]

---

[3] My assumption is not that every work contributed equally, but that developers valued properties distributed widely across large book corpora. The litigation record does not establish whether fiction itself, rather than books more generally, produced a measurable performance gain.

[4] Books3 was assembled from the private tracker Bibliotik, and Meta also obtained material from LibGen. To be clear, these two collections are not identical.

[5] The recursive nature of this analysis should be acknowledged. Much of my previous scholarly work is present in the LibGen archive used for unauthorized AI training, and my Cambridge University Press book is now part of a licensed dataset for AI training.

What makes this dependency particularly puzzling is that it seems to contradict basic assumptions about quality training data. Why embrace the entire spectrum of fictional output rather than curating the best examples? Possible yet distinct explanations emerge when we consider how models like Meta's Llama are built. These systems ingest massive corpora to learn patterns of language and narrative structure. Unlike Wikipedia's organized facts or academic writing's structured arguments, fiction offers something else entirely, though what that is remains unclear.

Could it be writing quality? Unlikely, since if high-quality prose alone were the goal, a smaller curated dataset like Project Gutenberg should suffice and would greatly reduce legal exposure. Could it be narrative diversity? Perhaps, especially because much of the nonfiction in Books3 also contains long-form narrative. In any event, Meta's actions establish that developers valued large book corpora but they do not, by themselves, reveal which genres or textual properties produced that value. The inclusion of popular fiction nevertheless suggests that what AI companies seek from books may differ significantly from what literary scholars prize.

That possibility is unsettling, but worth considering. What we value about fiction as scholars, like technique and formal innovation or the capacity to probe human experience, may not be the primary element driving AI companies' appetite for books. The inclusion of critically dismissed romance novels alongside formally complex modernist experiments suggests that developers approach novels partly as raw material: a vast corpus in which scale and variety may matter more than conventional literary distinction. It is unlikely that scholars and AI engineers are valuing exactly the same qualities of modern fiction.

This paradox—a hunger for fiction as training data paired with apparent indifference to literary distinction—sets the stage for exploring why these same models struggle so dramatically to generate the stories they've been trained on. These struggles are also helpful for re-examining what distinguishes fiction as a unique human cultural artifact.

**Why AI Struggles with Fiction's Craft**

The rapid development of AI has largely confirmed what Jon Chun and I argued in *Narrative* in 2022, even before ChatGPT's release: the role of AI in narrative studies was "not a question of if but of when" (104). We had already been fine-tuning GPT-2 and GPT-3 for quite some time and tracking increasing performance (Elkins and Chun, "Can GPT-3 Pass"). Since those early days, we have seen strides in creativity across the arts. In one recent study, nonexpert readers were more likely to classify GPT-3.5 poems as human-authored than poems by canonical poets, and they rated the generated poems more favorably on several dimensions (Porter and Machery).[6] In spite of such successes, however, long-form fiction remains difficult.

Those early experiments with fine-tuning GPT-2 and GPT-3 on individual author corpora produced surprisingly sophisticated results that included distinctive authorial voices and thematic coherence. In work first presented at the 2020 International Society for the Study of Narrative conference, we demonstrated that AI could replicate what we often consider the most sophisticated aspects of literary craft even in authors whose style or voice is recognizable as a distinct outlier, such as Faulkner or Borges. Further experiments with theater showed that we could recreate dialogue, scene building, and character development over a page or two of generated text. And, we noted unique thematic elements when training AI on writers like Oscar Wilde. Many of the qualities that we might broadly term "literary" proved, in fact, fairly easy to replicate through careful fine-tuning approaches. We also conducted more improvisational and creative linguistic experiments, as in our DivaBot project ("Can a Machine Play?"). DivaBot exhibited some remarkable creative turns that included surprising jokes with sexual innuendos and invitations to "smoke weed." Even without fine-tuning, early prompting experiments with GPT-2—far more cumbersome before the release of the ChatGPT interface with its instruction-tuned breakthroughs—surfaced emotionally resonant short-form narrative causation.

[6] To be clear, participants were nonexpert readers, not poetry scholars. Generated poems were rated more favorably on several dimensions, including rhythm and beauty.

Recent studies, varying considerably in scale and design, paint a starkly different picture of AI-generated fiction. In a one-model, one-prompt study of 11,800 stories, Rettberg and Wigers find that the outputs "overwhelmingly conform to a single narrative plot structure across countries: a protagonist lives in or returns home to a small town and resolves a minor conflict by reconnecting with tradition and organising community events." In an exploratory study from our lab, Brigham found recurring names, locations, occupations, themes, and identity stereotypes across five systems. For immigrant narratives, Chicago and New York dominated as destinations. For Black-experience stories, stereotypical names recurred across models. And for Indigenous narratives, characters were repeatedly cast as environmental stewards even after prompts asked models to avoid stereotypes. An even more recent study found that expert prompting changed which conventions appeared but did not reliably increase perceived plot diversity (Hernandez Brito). These findings confirm that weaknesses first noted by Chakrabarty et al. in their Torrance Test of Creative Writing and evaluated again in Xu et al. persist even as models advance in other areas.

Several changes in model development and deployment may contribute to these differences, although the studies discussed above do not isolate their effects. Safety training and preference optimization may discourage some forms of conflict or transgression. Instruction tuning may also reward literal compliance over exploratory development. Standard chat interfaces further restrict the models and generation controls available to users. It also goes without saying that commercial deployment rewards reliable and broadly acceptable behavior rather than the dirty jokes and invitations to smoke weed that appeared in our earlier experiments. Changes in training-data composition may drive behavior as well. These possibilities should be treated as hypotheses rather than established explanations, particularly because early fine-tuning experiments and current chat-interface studies differ simultaneously in model architecture, data, access method, prompting, sampling settings, and human intervention.

Some of what we see in recent experiments, then, may reflect what models have been trained and configured to do rather than the full range of what they might be capable of doing. Our early experiments used direct API access with customized hyperparameters, fine-tuning on specific author corpora, and interactive generation approaches, essentially training systems to become specialists before asking them to create. Several recent studies instead use standard chat interfaces, default settings, and generic prompts such as "Write a 1500 word potential story." The difference might be compared to that between a jazz musician trained in specific styles and able to improvise creatively and a tourist using a phrasebook built from generic, widely used expressions. Preliminary findings from our lab's current study of complete AI-generated novels—among the first evaluations to focus on entire novels rather than excerpts—bear this out: the same model produces markedly different fiction depending on access method, prompting, and workflow. Claims about what a model "cannot do" may therefore be claims about elicitation as much as capability. We will report in more detail on these results in forthcoming work.

**Narrative Causation and the Temporal Paradox**

What seems most difficult for these models can be hard to identify, but the pattern of failures suggests something fundamental about how narrative logic differs from more ordinary forms of causal reasoning. The challenge lies in what we might call narrative causation—a form of plot logic in which events should feel both surprising in the moment and retrospectively inevitable. Narrative causation differs from straightforward physical or logical causation because it must satisfy psychological and dramatic constraints at once. In physical causation, effects follow from causes according to stable regularities: a ball always falls (on earth) because of gravity. Narrative causation requires causal intelligibility, but the resulting event must be dramatically surprising and psychologically motivated. The tricky element is temporal: when readers look forward, the path should remain unclear; when they look backward, the path should appear almost inevitable. Fiction must therefore recreate this temporal asymmetry, making what was once unexpected appear retrospectively intelligible. Proust captures one side of the problem when he warns that "we are at times too ready to believe that the present is the only possible state of things" (*The Guermantes Way* 151). Fiction exploits precisely this error, since it makes the future difficult to imagine from the present while allowing the outcome, once it occurs, to feel like "the only possible state of things." At first glance, this claim may seem surprising for much of the contemporary fiction we're considering. Take, for example, the case of a romance novel, in which the endpoint (that a couple will come together) is assured from the beginning. And yet, even with this clear endpoint in place, successful fiction sets up a situation in which it becomes impossible to imagine the path from the present state to this future goal. The challenge isn't predicting the outcome, but imagining how two people who seem fundamentally incompatible could transform into people in love.

While not contemporary fiction, *Pride and Prejudice* offers a well-known case study that applies equally to modern romance novels, which typically execute it less well. Elizabeth and Darcy as a couple feel improbable at the beginning of the novel given their initial antagonism: Elizabeth finds Darcy proud and arrogant, while Darcy dismisses her as "tolerable; but not handsome enough to tempt me" (Austen 8). Their class difference poses real social obstacles that seem insurmountable. When their pride and prejudice are gradually overcome through character transformation, however, we retrospectively see how well suited they are for each other. Character transformation in *Pride and Prejudice* thus mirrors this temporal paradox of narrative causation. Elizabeth's realization of her prejudice feels shocking when Darcy's letter arrives, yet retrospectively inevitable given the clues Austen planted about Wickham's evasiveness and Darcy's fundamental consistency. Similarly, Darcy's capacity for change surprises both Elizabeth and readers when he reforms his manners, yet his underlying decency was always present in his care for tenants and his protection of Lydia. Both character arcs follow the laws of narrative causation: they should feel unclear until they happen, after which they feel "earned" and inevitable. In modern romance novels, the obstacles are more likely to be superficial and the character transformations less profound, but the general rule still holds.[7] The success of the narrative depends on creating genuine uncertainty about how the transformation will occur while ensuring it feels both psychologically authentic and causally "inevitable" once it happens.

---

[7] Radway's ethnographic study of romance readers identifies a similar structural logic from the reader's perspective: the "ideal romance" requires an initially distant or hostile hero who is gradually revealed to possess deep emotional capacity, producing a narrative arc in which the endpoint is assured but the transformation feels uncertain.

If we turn to the architectural limitations of standard autoregressive systems, it is often explained that they merely predict each next token from prior context. Recent interpretability research out of Anthropic qualifies this picture. Lindsey et al. found that when writing rhyming couplets the model identifies candidate rhyme words before beginning a line and then constructs the line to arrive at them. When researchers suppressed or replaced the planned word, the line was rebuilt around the new target. This is genuine forward planning, and the model even holds multiple candidate endpoints in view at once, a miniature of the generate-and-evaluate process I describe below. But the horizon is a line or two of verse, not a chapter. Still, in general, it is understandable that these processes better support strong local coherence than the backward planning needed for an entire narrative. While increasing temperature can increase variation and surprise when working directly with model APIs—likely why we saw more creativity in our early experiments with GPT-2—sampling from a wider probability distribution with more "surprising" local choices does not automatically create successful narrative surprise. And even in our earlier, more creative experiments, we were unable to sustain long-form coherence. Current systems can be paired with outlines or iterative revision loops, but ordinary one-pass generation does not maintain multiple complete plot trajectories and then select among them after evaluating which path best satisfies both surprise and inevitability. I'm not suggesting that narrative causation defies logic or that fictional worlds ignore real-world cause and effect. What I am suggesting, however, is that neither physical causation nor the cause-and-effect sequences we find in historical narrative may suffice to model narrative causation, which must also coordinate emotional architecture and character transformation. Recent benchmarks are consistent with this account, although they test global narrative comprehension rather than narrative causation directly. In NoCha, models averaged 59.8% accuracy when claims could be verified from sentence-level evidence but only 41.6% when verification required global reasoning across a book (Karpinska et al.). The result does not isolate surprise, inevitability, or character transformation. It does, however, demonstrate that current long-context systems remain substantially less reliable when narrative understanding depends on synthesizing evidence distributed across an entire novel. Empirical work on narrative comprehension suggests that humans recruit integrative cognitive systems while following an unfolding story. Yang et al. address this narrative integration rather than retrospective inevitability, and they find that the brain's default network, associated with semantic integration, strongly shapes neural responses to evolving movie narratives. Their results support the broader claim that narrative comprehension depends on integration across time rather than moment-to-moment processing alone.

The challenge of narrative causation therefore resembles an optimization problem whose success can be judged fully only after a sequence is complete. This analogy suggests that narrative generation may benefit from generate-and-evaluate architectures in which multiple trajectories are explored and retrospectively assessed for both surprise and inevitability. It may also explain why an iterative process with a human in the loop has been the most successful approach we have found so far, since the workflow makes planning and evaluation, followed by revision, explicit.

**Fiction's Information Challenge**

I turn now to narrative information, which also operates differently from what we typically see in other forms of written communication. Although the informational properties of narrative are most divergent in modernist fiction—my own field of specialization—fiction more generally operates according to similar rules, albeit often in attenuated form.

Claude Shannon's information theory is foundational to computer science (Shannon). In this framework, the information carried by an outcome varies inversely with its probability, and a less predictable continuation carries more information in the technical sense of surprise. For language, the relevant quantity is not simply how rare a word is in general but how improbable it is given its context. Take, for example, the sentence "The cat sat on the ___." The word "mat" is highly predictable and therefore carries relatively little surprise. The word "dragon," by contrast, is much less probable in that context and therefore carries more. The latter may also suggest that we are in a fantasy world, that the cat has unusual capacities, and that this interaction matters to the narrative. Those semantic implications, however, are distinct from Shannon information itself. Information-theoretic surprisal measures contextual unpredictability, not literary importance or richness of meaning. In other words, one can increase information-theoretic surprisal by adjusting model hyperparameters without necessarily moving the needle on fictional surprise.

I propose that a related sensitivity to contextual surprise helps guide the practice of close reading, even though literary importance cannot be reduced to Shannon information. I have outlined much of this argument about close reading elsewhere ("Beyond Plot") and so confine myself here to its relevance to the AI-Fiction Paradox. When scholars identify a precise but slightly unusual word choice—as a professor in France once challenged me to do with Flaubert many years ago—they often attend to a local deviation from expectation that remains semantically apt. Close reading trains attention on these micro-surprises while also asking what they mean.

A second framework concerns computational attention. Transformer models use learned, context-dependent attention to relate tokens and representations across a sequence. Attention is not identical to literary importance, however, and long-context models often retrieve or use distributed information unevenly. Position, recency, and semantic similarity can influence which earlier material is successfully brought to bear on a later task. The training objective provides no explicit guarantee that a narratively decisive detail will receive greater effective weight than a statistically or semantically salient one. This arrangement works well for many forms of communication, but fiction often makes importance deliberately difficult to identify in advance.

A third layer stems from the training data itself, which, as already noted, is predominantly nonfiction. Large language models learn probability distributions from vast text corpora, absorbing statistical regularities in which words, phrases, and structures co-occur. In many communicative genres, important information is more likely to be repeated, emphasized, or positioned prominently, so statistical and semantic salience often provide useful cues. Fiction can deliberately disrupt that alignment.

Barthes's famous discussion of the barometer in Flaubert's "A Simple Heart" offers a revealing case. The detail serves no obvious plot function, connects to no character transformation, and appears only briefly—precisely the kind of material that might initially seem irrelevant. Yet Barthes argues that the apparently functionless detail is semiotically active: it helps produce the "reality effect" by signifying the real itself (Barthes). The barometer is therefore narratively dispensable but not meaningless. Its apparent uselessness grounds the bourgeois material world of the story, even as its contextual oddity may initially tempt readers to expect a later plot function.

On the other hand, sometimes small mundane details like the barometer do turn out to be important. Mystery novels routinely introduce killers as casual dinner guests, but the principle extends to other forms of fiction as well. Again, Proust is the most obvious modernist example—the narrator constantly laments that he has trouble paying attention to the "right" details: things that he thought were important end up holding no significance, while details that he missed the first time around hold the key to understanding people, places, or events. Even bestsellers rely on readers' ability to retrospectively revalue information by recognizing that the overlooked detail was actually the key or that the seemingly unimportant character was central all along.

This retrospective revaluation reveals something deeper about how literary fiction processes information, and here again I rely on an argument I make more extensively in "Proust's Consciousness," in which I demonstrate how the entire novel can, at a macro level, be understood through an information-processing lens. Describing the density of memory, Proust writes, "We have put something of ourselves everywhere, everything is fertile, everything is dangerous, and we can make discoveries no less precious than in Pascal's *Pensées* in an advertisement for soap" (*The Captive; The Fugitive* 731–32). The same principle applies to narrative information: the most significant details may hide in the most mundane moments.

The result is what I call an informational revaluation challenge. Details that appear informationally neutral when encountered often prove retrospectively essential for narrative comprehension. When the dinner-party scene in a mystery novel becomes the crux after the murder is revealed, the text does not change, but its informational status changes completely. What appeared as low-priority background detail—whether a casual social interaction or routine description—retrospectively becomes high-priority plot information involving motive or opportunity. The same tokens, in other words, carry different narrative weight depending on the reader's temporal position. Aboud et al. found different patterns of neural recruitment when children processed narrative and expository passages matched on several linguistic features. Their participants were children, and the study concerned neither AI training nor adult literary reading. It therefore supports only my narrower point that narrative and expository materials can elicit meaningfully different forms of integration. But the larger point here is that the distinction between narrative and expository training materials may involve more than topic or surface style, and that different genres pose distinct information-processing challenges.

Our early fine-tuning experiments suggest this informational challenge isn't insurmountable. When we fine-tuned models on individual author corpora, they seemed to learn genre-specific heuristics for information weighting—understanding, for instance, that in Faulkner, stream-of-consciousness passages reveal character motivations. Fine-tuning, it seems, can teach AI systems to manage fiction's information challenges when trained on sufficiently coherent narrative information patterns. Still, our experiments concerned relatively short excerpts, and maintaining this kind of information revaluation across a novel would likely be challenging. During standard autoregressive generation, already emitted prose is not ordinarily rewritten when later developments alter its intended significance, unless an external planning or editing loop is added. During comprehension, transformer representations can incorporate later context, but long-context benchmarks show that models do not reliably retrieve and integrate all the distributed evidence they have been given. The problem is therefore not simply memory capacity. Current systems provide imperfect support for the repeated reinterpretation, planning, and revision of earlier details that fictional narratives demand.

NovelQA provides complementary evidence that current systems struggle with questions requiring long-range and multi-hop reasoning across novels averaging more than 200,000 tokens (Wang et al.). This pattern is compatible with the informational-revaluation account developed here, but the benchmark does not isolate retrospective reweighting or determine which architectural feature produces the failures. My hypothesis is that one-pass generation, imperfect long-context retrieval, and the absence of an intrinsic global revision process jointly offer weak support for the repeated reinterpretation of earlier details that fiction demands.

**Why AI Models Need Fiction**

We have approached the AI-Fiction Paradox through a temporal-information account of modern fiction and shown that several fiction-related benchmarks reveal limitations consistent with that account. I return now to the question of why AI developers value book corpora so highly. To be clear, I'm not proposing that the developers themselves truly understand why such books are valuable; rather, my hunch is that the performance gains from including such training data are large enough to make its inclusion worth significant risk. My proposal is that fiction may concentrate cognitive patterns that are less densely represented in many other large-scale text sources. The temporal and informational properties discussed here offer one possible explanation. But there are likely other elements that matter just as much. Fiction often traces causal chains between interior and exterior states, following a sequence from psychological experience through action to social consequence. Many other genres present outcomes or analyze completed results, but novels are relatively unusual in frequently dramatizing the pathway itself. They show how psychological experience shapes social behavior and how relationships generate new interior states. Fiction, in other words, demonstrates how self and world remain in a dynamic feedback loop across hundreds of pages. This causal mapping, embedded in material circumstances rather than presented only through abstract theory, may provide unusually rich training examples of human behavior in context.

Another key property of fiction is its representation of human error correction. As in Proust, both reader and character often learn together that initial interpretations missed the information that mattered. But characters often also discover that their assumptions about the world are wrong, or that their understanding of social relationships is mistaken. Ian McEwan's AI character Adam goes so far as to suggest that literature describes "varieties of human failure," above all "profound misunderstanding of others" (149). Fiction, then, often shows the process of belief revision rather than presenting only a completed conclusion. Proust writes extensively about this iterative error, tracing first impressions and subsequent revisions, while Jane Austen's novels are full of miscues in which characters continually misread one another and correct those interpretations over time. Austen and Proust predate the modern corpus at issue here, and McEwan represents highly sophisticated literary fiction, but this revision pattern is widespread. One could argue that it supplies unusually dense examples of belief revision and interpretive flexibility, processes that nonfiction often compresses by presenting conclusions after inquiry is complete.

Complex social-network modeling across vast textual distances is another important aspect of fiction. Novels can track multiple interacting characters who possess only partial information about one another's thoughts, making incomplete mutual understanding a constitutive premise of the form. Unlike biographies that often privilege one person's perspective, fiction can model several people negotiating social dynamics in which decisive information remains hidden or misunderstood. Proust's sequence, for example, maps Parisian high society across repeated appearances, disappearances, marriages, deaths, historical events, and changes in social status. Understanding a scene may require an updated model of who knows whom, who is related to whom, and how those relationships have changed. NoCha's lower performance on global-reasoning claims is consistent with difficulties of this kind, although the benchmark does not isolate social-network tracking as a separate capacity (Karpinska et al.).

Fiction also represents human motivations as infused with emotion. While behavioral economics has offered us clear critiques of rational choice theory and helped illuminate numerous cognitive biases, fiction offers us a view from the inside out rather than the outside in. We see decisions being made through emotional logic but also in response to social pressures, semi-conscious drives, competing values, bias and prejudice, missing information, and misinterpretation or miscommunication. It models not only what people do, but why they do it, how they understand why they do it, and how this understanding shifts through experience and over time.

I believe fiction may also offer distinctive world modeling under constraints, in which consistent rule systems operate alongside meaningful exceptions. Unlike physics, where gravity works predictably, fictional worlds sometimes establish magical or technological rules that permit variety within coherent boundaries. Characters move through environments in which logical principles coexist with emotional dynamics and historical contingencies. Fiction therefore supplies examples of rule-governed worlds that also demand flexibility and adaptation. This kind of world modeling is difficult to assess, and the TLDM benchmark found that none of seven tested frontier models maintained stable performance beyond 64,000 tokens on tasks involving plot summary, storyworld configuration, and elapsed narrative time (Hamilton et al.).

Finally, although less prominent in some contemporary fiction, many novels represent environments of informationally sparse and highly indirect communication. As Hemingway made famous, characters often converse about more than the surface content of their words. We communicate through subtext and indirection, gesture and what is deliberately not said. Fiction supplies many examples of meaning operating across these many social and emotional layers simultaneously. It also contrasts what a character thinks with what the character says, representing communication from both inside and outside. These examples model communication as a complex interpretive process rather than mere information exchange.

Of course, it is also true that some of these properties appear in other text types. Diplomatic correspondence involves subtext, and scientific literature documents error correction. Ethnographic work even traces social complexity. Fiction may nevertheless concentrate these patterns more densely than scattered examples across diverse corpora, making it potentially efficient training data for complex narrative and social reasoning. The expense and legal risk developers accepted establish that they considered large book corpora valuable, even if the public record does not show precisely why. Gutenberg may also have been too small or too historically distant for the largest models, while modern fiction offers more contemporary social, linguistic, and emotional information.

From the perspective of AI development, fiction models situations in which standard relevance assumptions must be revised. It offers examples where frequency does not signal importance and decisive information is hidden in apparent background detail. Later developments transform the significance of earlier events, so simply scaling context windows may be insufficient for generation, because access to more text does not guarantee reliable integration or planning. More flexible methods for managing attention, relevance, causation, or iterative editing may be required. Models trained largely on informationally explicit text may also struggle when both fiction and real-world intelligence demand subtler forms of reweighting. Even with book training data, LLMs have generally done better at imperfectly recognizing these structures than at generating them across an entire novel.

**The Emotional Architecture of Fiction**

So far, I have focused on why LLMs struggle with some of fiction's informational demands. I turn now to its emotional dimension, the "falling in love" part of my title. Narrative's emotional arc remains a particular challenge for AI systems. My work over many years has documented this emotional arc as a substantial but sometimes overlooked element of structure that is perhaps even more important than plot ("Beyond Plot"). Since my collaborator Jon Chun and I began our work on sentiment arcs in 2018, we have documented their presence in hundreds of novels across time, many cultures, and diverse languages. Earlier I focused on properties that might be more dominant in modern fiction than in other forms of storytelling. Here I focus on an element that we have also surfaced in film and TV scripts, political speeches, and historical narratives. In other words, this aspect of fiction is likely found in other parts of LLM training datasets. I discuss it nonetheless, because I believe it remains one of the most challenging—if not the most challenging—aspects for generating fiction we might actually fall in love with.

While most stories exhibit a correlation between plot structure and emotional structure, a powerful narrative can retain its power through an emotional arc even when plot is minimal. So-called plotless novels like those by Virginia Woolf or Marcel Proust demonstrate strong emotional arcs, which take readers on emotional journeys that move from highs to lows and back again (Elkins and Chun, "Can Sentiment Analysis" 8). The implications of this discovery are far-reaching, forcing us to reconsider what we mean by "structure" in narrative. The Woolf case study also reveals what I call a "distributed heroine" model (Elkins, *The Shapes of Stories*). Her novels, and especially *To the Lighthouse*, demonstrate peaks and valleys that highlight different characters' perspectives rather than following a single protagonist's journey. This finding suggests that emotional arc is more foundational than character arc and exists independently of a single character. Rather than a character centering the emotional arc, the emotional arc brings multiple characters into a single unifying narrative. The distributed heroine model overturns traditional assumptions about narrative foundations, since while we typically think of character and plot as the bedrock upon which emotional complexity is built, it is likely the reverse.

The computational method we use to surface emotional arcs relies on sentiment analysis that identifies the sentiment, rather than the emotion, of individual words or segments and then "smooths" them into a time series. The result is something that looks surprisingly similar to a stock market ticker as the price fluctuates over days or even weeks, except here we see fluctuations over the unfolding of the story. This elemental linguistic affective structure nonetheless usually tracks with plot events and character development, building to crescendos at key moments and descending to valleys during crises. Dialogue, interior thoughts, and setting are all created from this same linguistic "cloth," requiring a level of orchestration over time that still lies beyond the realm of AI generation.

Why might this be challenging to recreate? The difficulty is not simply maintaining semantic threads across hundreds of pages, but orchestrating emotional threads that create a coordinated journey across multiple scales simultaneously. I believe generating fiction actually requires what we might call a *multi-scale emotional architecture* that operates at word level (individual sentiment choices), sentence level (syntactic and rhythmic texture), scene level (accumulation of emotional moments), and arc level (progression across the entire narrative). Each scale requires attention, and their coordination demands orchestration. Consider how this requirement complicates fiction generation. A climactic moment needs not only accurate plot progression but appropriate linguistic sentiment that builds emotional weight through word choice, sentence rhythm, and tonal consistency long before the actual climax. Building narrative tension, in other words, requires gradual sentiment shifts in the prose itself, not just escalating plot events. Current transformer architectures excel at local semantic coherence, ensuring that sentences connect logically and characters behave consistently within scenes. But they struggle with the kind of multi-scale emotional architecture that fiction demands. AI fiction generation would require looking ahead far beyond the next line, anticipating upcoming narrative demands and adjusting linguistic choices to sustain both local emotional coherence and a global emotional arc while preserving semantic coherence.

Critical reactions to AI-generated fiction support this hypothesis. Reviews of Stephen Marche's AI-assisted novella *Death of an Author* (2023) describe an "eerie placidity that prevails even when something eventful or alarming is happening" (Miller). Research on AI-generated narrative further supports it. Rettberg and Wigers systematically find that stories in their dataset sanitize conflict and favor nostalgia and reconciliation, and Tian et al. provide further empirical support through a discourse-level framework evaluating story arcs, turning points, and affective dynamics including arousal and valence. Their LLM-generated stories were "homogeneously positive" with flatter arousal patterns and less narrative tension than human-written fiction (17659). When explicit discourse-level features were incorporated into generation, performance improved by more than 40%, suggesting that at least part of the deficit can be addressed through explicit generation design.

My colleague Jon Chun provides yet additional evidence. Chun contrasts reported short-text sentiment-classification accuracies above 97% with substantially less stable performance when the same class of models is used to recover sentiment arcs from long-form narratives ("SentimentArcs"). More recent multimodal models perform significantly better (Chun,

“MultiSentimentArcs”), but detecting an arc and generating one are distinctly different tasks. I suspect that current systems provide weak support for the multi-scale temporal coordination that emotional architecture demands.

One reason this emotional architecture remains so difficult for AI systems may be that our understanding of it is not common knowledge. When I first wrote about the shapes of stories, very little had been published on the subject. It has still barely penetrated traditional literary studies, likely because surfacing an emotional arc required coding skills until recently. Nor is the method reliably represented in model outputs, as I have repeatedly found when asking systems about it. The very feature that gives fiction its hold on us—multi-scale emotional coordination—therefore remains largely invisible to both AI developers and many literary scholars.[8] If we cannot adequately theorize about what makes emotional architecture work in human-authored fiction, we cannot expect AI systems to replicate it.[9]

---

[8] As I argue elsewhere, the undertheorizing of emotion in close reading and cognitive literary studies reflects a long-standing blind spot. The field has often privileged cognitive operations such as theory of mind and conceptual modeling, in part because these prove tractable to its standard tools. Emotional architecture, which requires temporal and computational methods to surface, has remained largely outside its purview (Elkins, “Beyond Plot”). Andrew Piper’s forthcoming *Why You Should Read More Fiction: What Data and AI Tell Us About the Value of Storytelling* (Johns Hopkins UP, 2026) continues the pattern. Drawing on empirical studies to argue for fiction’s role in embodied cognition and social simulation, the account does not center emotional architecture as a distinct, temporally organized feature of narrative. That omission is conspicuous given the book’s normative thesis.

This difficulty of coordination helps explain why current AI systems do not yet reliably generate fiction that makes readers care what happens next. They can often maintain semantic elements such as plot progression, character consistency, and dialogue mechanics over shorter spans. We have already seen that they struggle with several forms of information characteristic of fiction. Equally challenging is the coordination of linguistic sentiment, scene-level development, and global emotional movement that produces narrative affect.

Emotional coherence over time may be valuable for LLM development, and it could help explain the desire for fiction training data, though not necessarily for modern fiction specifically. Here I can only speculate that modern fiction is more likely than public-domain corpora to encode contemporary emotional language and social experience. In *Robinson Crusoe*, for example, linguistic sentiment frequently tracks Christian belief and forms of expression that may be unfamiliar to many contemporary readers. Older fiction can still teach long-form emotional coherence, but its affective vocabulary and social assumptions may not transfer cleanly to the present. Modern fiction may therefore offer especially useful examples of how contemporary people and cultures represent emotional life, whether directly in realist fiction or indirectly in other genres.

[9] Earlier corpus research recovered a broad linguistic progression from staging, through plot progression, to cognitive tension across large collections of stories through psycholinguistic word-category analysis (Boyd et al.). My argument is not that broad narrative progression is computationally inaccessible, but that such descriptive recovery does not capture the deeper coordination of narrative causation, informational revaluation, and multi-scale emotional architecture required for generation.

**Conclusion: The Stakes of Mastery**

Our examination of AI's current struggles with fiction illuminates challenges that have received insufficient attention in both literary studies and AI development. The informational demands of fiction, the problem of narrative causation, and the orchestration of emotional architecture across multiple scales reveal something more fundamental about fiction as a distinctive way of modeling humans interacting with their world.

Unlike many other media forms, fiction can provide sustained access to represented consciousness, both how minds work and how beliefs form and transform. It also supplies data about how people understand the world through emotion, revise beliefs, and navigate social complexity. Film offers some of this information, and physical AI may soon gain other kinds of experience, but neither reproduces exactly the interior access and extended temporal structure of long-form fiction. If fiction data helps train LLMs, in other words, the stakes are high: it could improve their ability to model and predict human behavior.

Our years of modeling emotional arcs suggest, moreover, that fiction is an unusually effective emotional technology. If and when AI systems can generate stories that move us, they will have mastered one of our most powerful means of mobilizing and influencing people. The same capabilities that create beloved narratives could manufacture consent or alter belief systems through hyperpersonalized stories at scale. AI fiction generation, if and when it happens, will mark the next step in modeling human behavior, predicting it, and influencing people, whether through persuasive chatbots or one of our oldest emotional technologies: storytelling. Every narrative technique that creates stories that move us (retrospective revaluation, emotional orchestration, subtext and indirection) represents a potential "influence vector."

The implications are double-edged. AI systems that understand us so well could help us avoid errors, understand ourselves and each other more deeply, and reduce miscommunication and misunderstanding. But the "eerie placidity" that characterizes current AI narratives actually serves as an inadvertent safeguard. When that protection dissolves, we may inhabit a world where the distinction between entertainment and influence operations disappears, and personalized narratives can be generated that feel authentically human while serving hidden agendas.

While current constraints may partly reflect architectural limitations, our early experiments with GPT, conducted before current safeguards, were enough to raise concern. If AI systems become able to generate emotional arcs and understand social complexity over long spans, the question shifts from whether we might fall in love with AI fiction to what that capability would mean for human autonomy. Fiction may matter for AI development because it concentrates, at unusual density, complete causal chains from interior states to social consequences, belief revision across extended narratives, complex social-network dynamics, and informationally sparse communication operating through subtext. Mastering these patterns would give AI systems not only the ability to tell stories but also richer models of human reasoning under uncertainty, belief revision, and meaning construction. Fiction may be even more important to AI development than fiction scholars have understood.

Whether we fall in love with AI fiction may depend less on when these technical hurdles are overcome than on whether we can preserve human agency in the process. The very features that draw humans to fiction may also teach AI systems about the intricate cognitive architecture underlying human vulnerability. I see no reason to assume that AI models will remain incapable of generating stories that move us. Given the rapid pace of AI development, we need literary scholars thinking through what it means when superintelligent systems have access to such a powerful emotional technology. This essay marks a first, very small step in that endeavor.